\documentclass[10pt,twocolumn,letterpaper]{article}
\usepackage[accsupp]{axessibility}
\usepackage{iccv}
\usepackage{times}
\usepackage{epsfig}
\usepackage{graphicx}
\usepackage{amsmath}
\usepackage{amssymb}
\usepackage{algorithm}
\usepackage{algorithmic}
\usepackage{multirow}
\usepackage{makecell}
\usepackage{colortbl}
\usepackage{tabularx}
\usepackage{color}
\usepackage{booktabs}
\usepackage{multirow}

\newcommand\clearrow{\global\let\rowmac\relax}
\clearrow

\usepackage[colorlinks, breaklinks=true,bookmarks=false]{hyperref}

\iccvfinalcopy 

\def\iccvPaperID{****} 

\ificcvfinal\fi

\begin{document}

\title{BN-NAS: Neural Architecture Search with Batch Normalization}

\author{Boyu Chen\textsuperscript{1}\thanks{Equal contribution}, Peixia Li\textsuperscript{1}\footnotemark[1], Baopu Li\textsuperscript{2}\thanks{Corresponding author},  Chen Lin\textsuperscript{3}, Chuming Li\textsuperscript{1,4}, 
 Ming Sun\textsuperscript{4}, Junjie Yan\textsuperscript{4}, Wanli Ouyang\textsuperscript{1}\footnotemark[2] \\
\textsuperscript{1} The University of Sydney,
\textsuperscript{2} BAIDU USA LLC,  \\
\textsuperscript{3} University of Oxford,
\textsuperscript{4} SenseTime Group Limited \\
}

\maketitle
\ificcvfinal\thispagestyle{empty}\fi

\begin{abstract}
We present BN-NAS, neural architecture search with Batch Normalization (BN-NAS), to accelerate neural architecture search (NAS). BN-NAS can significantly reduce the time required by model training and evaluation in NAS.
Specifically, for fast evaluation, we propose a BN-based indicator for predicting subnet performance at a very early training stage. The BN-based indicator further facilitates us to improve the training efficiency by only training the BN parameters during the supernet training. This is based on our observation that training the whole supernet is not necessary while training only BN parameters accelerates network convergence for network architecture search. 
Extensive experiments show that our method can significantly shorten the time of training supernet by more than 10 times and shorten the time of evaluating subnets by more than 600,000 times without losing accuracy. 	The source codes are available at \href{https://github.com/bychen515/BNNAS}{https://github.com/bychen515/BNNAS}.
\end{abstract}

\section{Introduction}
Neural architecture search (NAS), which aims to find the optimal network architecture automatically, has significantly improved the network performance in many computer vision tasks, such as image classification~\cite{NasNet-CVPR18-Zoph, SPOS-ECCV20-Guo, li2020improving, ci2020evolving, chen2021glit}, object detection~\cite{CRNAS-ICLR19-Liang, DetNAS-NIPS19-Chen, liu2021inception}, semantic segmentation ~\cite{DeepLabv3+NIPS18-Chen, AutoDeepLab-CVPR2019-Liu}, etc. However, a successful NAS method usually means training and evaluating thousands of models, which takes up to thousands of GPU days~\cite{NasNet-CVPR18-Zoph, AmoebaNet-AAAI19-Real}. The huge searching budget makes NAS hard to be applied widely. 

To overcome the above issue, one-shot methods~\cite{ENAS-ICML18-Pham, SPOS-ECCV20-Guo}, have been proposed to reduce the computational cost based on the weight-sharing technique, reducing the search cost from thousands of GPU days to tens of GPU days.
These methods construct a supernet that includes all candidate network architectures.  With the constructed supernet, one-shot methods consist of three stages: supernet training, subnet searching and subnet retraining. 

In the supernet training stage, the supernet is trained by back-propagation. 
In the subnet searching stage, subnets are sampled from supernet and treated as the candidate architectures. The sampled subnets are evaluated on validation data, from which the top-5 subnets with the highest accuracy on validation data are selected in SPOS. 
The selected subnets are then retrained from random initialization in the subnet retraining stage. The primary benefit of one-shot methods is that the subnets can inherit the weights of supernet to reduce the computational burden significantly in the searching stage. However, the process of training supernet hundreds of epochs and evaluating thousands of subnets is still time-consuming, leading to tens of GPU days cost.

\begin{figure}[t]
	\centering
	\includegraphics[width=1\linewidth]{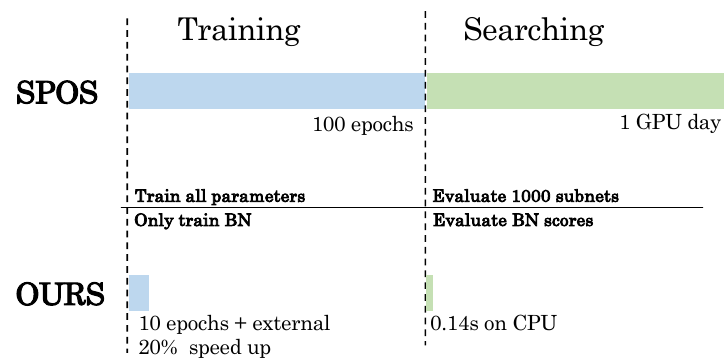}\\
	\caption{Designs and computational cost of SPOS and Our BN-NAS. 
	Compared with SPOS, our proposed BN-NAS can accelerate the one-shot methods in two stages: training supernet more than ten times faster, and searching subnets more than 600,000 times faster. The key to the speed-up is the BN-based indicator, which saves the searching cost and facilitates the training only BN paramters with much fewer epochs. SPOS needs 11 GPU hours in total. Ours needs only 0.8 GPU hours.}

	\label{fig:motivation}
	\vspace{-12pt}
\end{figure}

In this paper, we identify the parameters learned at the Batch Normalization (BN) layer as the key to significantly reduce the excessive time required by one-shot methods in training and searching stages.
In searching stage, the motivation is that BN parameter is a very light-weight measure for the importance of operations and subnets.
Existing one-shot methods evaluate thousands of subnets on validation data. Although the searching process efficiency has been improved, the large computation required for these thousands of subnets is still a burden. 
It is widely accepted that the BN parameter of a channel reflects the importance of the channel~\cite{NetworkSliming-ICCV17-Liu, OP-aware-pruning-arxiv20-Kang}. Hence, channels with smaller BN parameters are considered as less important and pruning these channels will cause a small influence on the whole deep network~\cite{NetworkSliming-ICCV17-Liu}.
Therefore, it is natural to accumulate the BN parameters from multiple channels to measure the importance of candidate operations and even the whole subnet.
Based on this observation, we propose a novel BN-based indicator to measure the importance of operations as well as subnets, which significantly reduces the searching cost from about 1 GPU day for SPOS to 0.14s on CPU for ours in the searching stage, as shown in the column for `searching' in Fig.~\ref{fig:motivation}.

The BN-indicator further motivates us to only train the BN parameters of supernet in the supernet training stage.
To train a supernet, it is a general practice to train all parameters, \ie, parameters of convolutional layers, fully connected layers, and BN layers. Yet training BN layers only is not groundless.
Frankle~\etal~\cite{TrainBN-arxiv20-Frankle} find that networks only training BN parameters with other randomly initialized parameters fixed still have a certain capacity. 
During the supernet training stage, our BN-NAS only trains BN parameter but does not train the other parameters such as convolutional or fully-connected layers for two reasons: 1) the network can encode the knowledge from training data through training only a part of parameters, as found in~\cite{TrainBN-arxiv20-Frankle}; 2) we focus on using BN parameters as the indicator for searching instead of network accuracy. We empirically find that training only BN parameters helps BN parameters become stable at earlier training epochs. 
Besides, there is an extra training speedup benefit from only training BN parameters. Based on the observations above, we propose a new BN-NAS. The BN-NAS train supernet with much fewer training epochs, and search subnets using the novel BN-based indicator for much faster speed.

To summarize, the main contributions are as follows:
\begin{itemize}
	\setlength{\itemsep}{3pt}
	\setlength{\parsep}{3pt}
	\setlength{\parskip}{3pt}

	\item We propose a BN-based indicator for evaluating network architectures, which can significantly shorten the time required by one-shot NAS methods for searching candidate network architectures.

	\item 
	We only train the BN parameters of the supernet and significantly reduce the number of epochs required for training the supernet, which is based on the use of BN-based indicator when evaluating network architectures. Training BN parameters only and reducing the training epochs could have adverse effects on the network architecture searching stage. However, with our BN-based indicator for searching, the adverse effect is overcome.

	\item Extensive experiments demonstrate that our method can significantly improve the speed of NAS in the training stage (more than 10X, for example, from 100 to 10, with an external 20\% speed up for SPOS as shown in Fig.~\ref{fig:motivation}) and searching stage (more than 600000X) without losing accuracy.

\end{itemize}


\begin{figure*}[!t]
	\centering
	\includegraphics[width=1\linewidth]{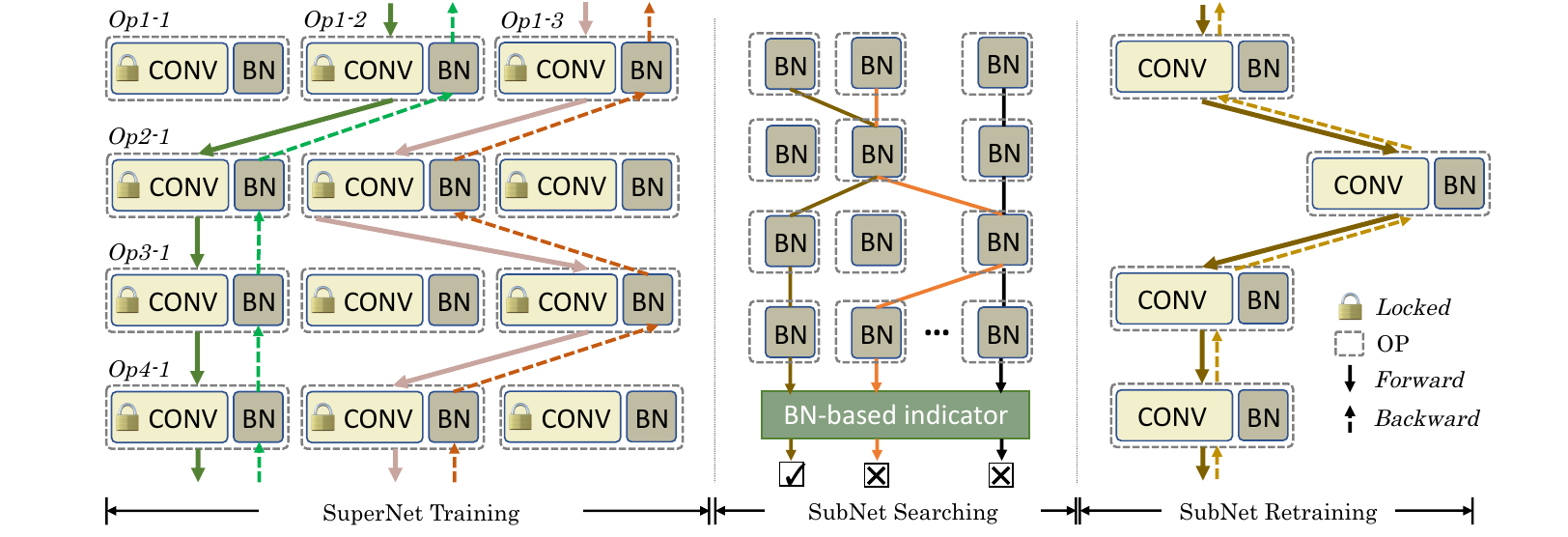}\\
	\caption{Overview of the proposed framework. We follow the three stages in one-shot methods. In supernet Training, we fix the convolution parameters and only train BN parameters for few epochs. In a iteration of supernet training, only a single path is sampled from the supernet for forward propagation, e.g. following the green solid arrows, and back-propagation, e.g. green dashed arrows. In Subnet Searching, we search the subnets (lines of the same color is a subnet) with proposed BN-based indicator. In Subnet Retraining, we train best subnet from scratch.}
	\label{fig:framework}
	\vspace{-16pt}
	
\end{figure*}

\section{Related Works}
\subsection{Reinforcement Learning and Evolutionary Algorithm for NAS }
NAS methods are proposed for automatic network architecture designing. Early methods utilize reinforcement learning (RL)~\cite{NasNet-CVPR18-Zoph, RLnas-ICML17-BelloZVL} or Evolutionary algorithm(EA) ~\cite{AmoebaNet-AAAI19-Real} to generate network architecture samples. The generated network samples are evaluated on validation dataset and their accuracies are treated as rewards to guide the RL and EA to generate better architecture samples. 
Zhou \etal~\cite{EcoNAS-CVPR20-Zhou} propose an optimal proxy for the Economical Neural Architecture Search. However, the sampling and training processes are still time-consuming, making it difficult for NAS to be deployed on large-scale datasets such as ImageNet~\cite{DengDSLL0-CVPR09-IMAGE}.

\subsection{Weight-sharing NAS}
To overcome the time-consuming problem of RL and EA, methods based on weight-sharing mechanism are proposed. These methods adopt the supernet constructed by all candidates subnets and divide the NAS process into three stages. Based on the difference in the training and searching stage, these methods can be divided into one-shot methods and differentiable methods.
{\flushleft {\bf{One-Shot Methods. }}} 
One-shot methods construct the supernet with candidates subnets directly and train the supernet based on sampled subnets for hundreds of epochs. After supernet training, thousands of subnets are sampled and evaluated on the validation set to find the optimal subnet architecture based on the validation accuracy. Since the search space is enormous, EA algorithm is adopted to generate subnets to be evaluated.
Most one-shot methods focus on subnets sampling during the training.
~\cite{SPOS-ECCV20-Guo} constructs the supernet and then trains the supernet through single-path random sampling. 
Based on~\cite{SPOS-ECCV20-Guo}, ~\cite{FairNas-arxiv19-Liang} proposes a fair sampling method to alleviate supernet bias and improve the evaluation capacity. ~\cite{greedyNAS-cvpr20-you} proposes a sampling pool and samples subnets in the pool during the supernet training, improving the training efficiency. Unlike the above methods, we only train the BN parameters in supernet which are based on different sampling policies for much fewer epochs.  Besides, we evaluate the subnets through our proposed BN-based indicator instead of evaluating subnets on validation set, accelerating the searching stage significantly.
{\flushleft {\bf{Differentiable Methods. }}} 
Different from one-shot methods, differentiable methods construct the supernet with additional architecture parameters. During the supernet training, the subnet sampling is controlled by architecture parameters which are trained alternatively with subnet parameters. After the supernet training, the optimal architecture is selected according to the magnitudes of the architecture parameters.
~\cite{Darts-ICLR19-Liu} treats the architecture parameters like the weight for the subnet output and updates the architecture parameters by back-propagation. ~\cite{Proxyless-ICLR19-Cai} binarizes the architecture parameters to save the GPU memory usage during the supernet training. ~\cite{SNAS-ICLR19-Xie} introduces Gumbel random variables to train the subnet and architecture parameters directly.
However, the architecture parameters bring the training tendency to certain operations during training, especially for skip connection.  Compared with these methods, our method does not need external parameters and can ensure fairness among all candidates during training.

\section{Method}

\subsection{Preliminary}
Since our approach is based on the One-shot NAS method~\cite{SPOS-ECCV20-Guo, FairNas-arxiv19-Liang} and the Batch Normalization Layer~\cite{BN-ICML15-Ioffe}. A brief introduction of them is provided in this subsection.

\vspace{-2mm}
\subsubsection{One Shot NAS }
In One-Shot (e.g. SPOS) methods, a supernet $\mathcal{N}$ with weights $\mathcal{W}$ is constructed by all candidate operations forms the search space $\mathcal{A}$. The whole pipeline of these methods can be divided into three stages, \ie supernet Training, Subnet Searching, and Subnet Retraining.

\vspace{-2mm}
{\flushleft { {{\bf{Search Space.}}}}}  The supernet architecture is constructed by a series of candidate operations as shown in Fig. \ref{fig:choice}. A layer contains multiple ($N$) candidate operations. Every candidate operation in the layer follows the repeating structure of `Conv-BN-ReLU' with different kernel sizes and expansion ratios (number of channels).

{\flushleft {{\bf{Supernet Training }}}}
The supernet $\mathcal{N}$ is trained by sampling a single-path architecture $a \in \mathcal{A}$ based on a sampling policy at each iteration.   
In the single path architecture search method,  only a single candidate operation in each layer is activated. Then the weights of the sampled architecture, denoted by  $\mathcal{W}_a$, are optimized by normal network training, \ie back-propagation. 

Since the accuracy of subnet with weights inherited from the supernet should be highly predictive on the validation set, the supernet training often requires hundreds of epochs.

{\flushleft { {\bf{Subnet Searching }}}}
After training the supernet, the next step is to find the optimal architecture with the best performance. In SPOS, the accuracy on validation set is used for evaluating the subnet performance. 
The optimal subnet is selected according to the subnet accuracy on the validation set.
To get a reliable searching result, thousands of subnets are needed to be evaluated.

\vspace{-1mm}
{\flushleft {{\bf{Subnet Retraining }}}}
In the retraining stage, the $K$ subnets found at the subnet searching stage with the highest accuracy are retrained. They are then evaluated on the validation set and the subnet with the highest accuracy is chosen as the final optimal subnet.

\vspace{-2mm}
\subsubsection{Batch Normalization Layer }
Batch Normalization (BN) layer has been used in network pruning ~\cite{NetworkSliming-ICCV17-Liu, EBticket-arxiv19-You,OP-aware-pruning-arxiv20-Kang}, which is a good evaluation of channel importance. Given the input $x^{in}$ of BN layer, the output $x^{out}$ is calculated through:
\begin{equation}\label{fuc:bn}
	\begin{aligned}
		z=\frac{x^{\mathrm{in}}-\hat{\mu}}{\sqrt{\hat{\sigma}^{2}+\epsilon}}, \\
		x^{\text {out }}=\gamma \cdot z+\beta,
	\end{aligned}
\end{equation}
where $\epsilon$ is a small positive value for numerical stability, $\hat{\mu} \equiv \mathrm{E}\left[x^{\mathrm{in}}\right]$ and $\hat{\sigma}^{2} \equiv \operatorname{Var}\left[x^{\mathrm{in}}\right]$ are means and variations calculated across mini-batches. The scaling parameter $\gamma$ and bias parameter $\beta$ are learnable parameters in BN layers to affine the normalized features $z$.

\subsection{Algorithm Overview}
 The pipeline of our proposed NAS method is shown as Fig.~\ref{fig:framework}. We follow the three stages in one-shot methods, \ie supernet training, subnet searching, and subnet retraining. In supernet training stage, the  supernet containing all candidate operations are randomly initialized. Only BN layer parameters are updated through standard forward-backward training, while the other parameters of the supernet are fixed after initialization (Section \ref{sec:only bn}). In subnet searching stage, subnets are sampled and evaluated based on our BN indicator (Section \ref{sec:bn indicator}). In the subnet retraining stage, the best subnet chosen in the subnet searching stage is retrained.

In the following, we start from the second stage (Subnet Searching). The order of the following description is consistent with the order of our exploration in this direction.

\subsection{Subnet Searching with BN Indicator}\label{sec:bn indicator}

Given the trained supernet,  we need to evaluate the performance of sampled subnets in the Optimal Subnet Searching stage. We utilize BN parameters to evaluate the performance of candidate operations.

\textbf{Change of Denotation for BN layer.}
Different from channel pruning, we focus on the operation-level outputs instead of channel-level. Take the $c$-th channel of activated operation $o_l$ output ($C$ channels in total) in layer $l$ as the example, the denotation for the BN layer need to be changed correspondingly.
Eqn.~(\ref{fuc:bn}) can be rewritten as follows for the $c$-th channel of operation $o_l$:  
\begin{equation}\label{fuc:bn-block}
	\begin{aligned}
		z_c=\frac{x^{\mathrm{in}}_c-\hat{\mu}_c}{\sqrt{\hat{\sigma}_c^{2}+\epsilon}}, \\
		x^{\text {out }}_c={\gamma}_c \cdot z_c+{\beta}_c,
	\end{aligned}
\end{equation}
where symbol with subscript $c$ represents the parameters in the $c$-th channel as the definition in Eqn.~(\ref{fuc:bn}). Assume the normalized features in $z$ follow the normal distribution $N(0, 1)$, a smaller scaling parameter $\gamma$ means a smaller magnitude of BN layers output $x^{\mathrm{out}}$. Since the output of a channel with smaller magnitudes contribute less for whole network~\cite{NetworkSliming-ICCV17-Liu},  we can treat the scaling parameter $\gamma$ as the importance of the channel.

\begin{figure}[h]
	\centering
	\includegraphics[width=1\linewidth]{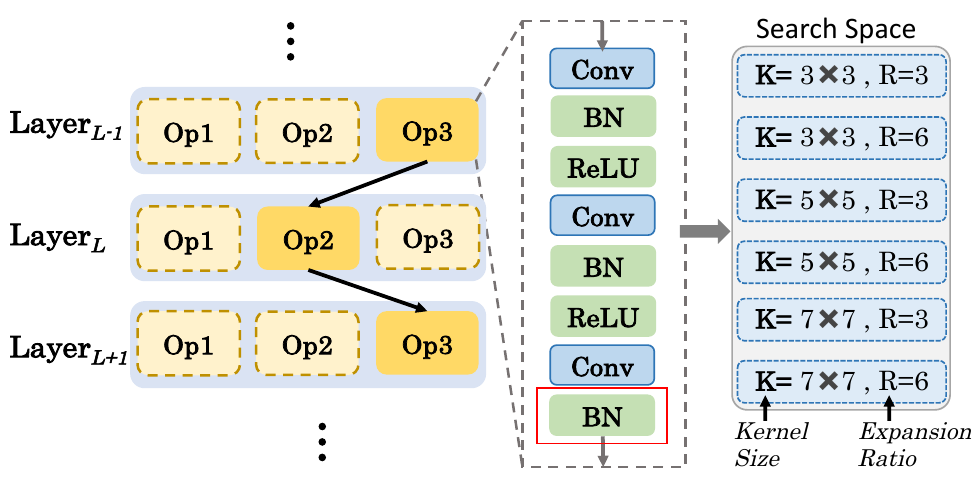}\\
	\caption{The single-path based search space. Only one operation is activated in each layer during network forwarding. We focus on the most popular search space for mobile setting network searching~\cite{EffNet-ICML19-Tan, FBNet-CVPR19-Wu} as the shown in right. The candidate operation consists of a series of Conv-BN-ReLU and ends with Batch Normalization(BN). We search the optimal kernel size and expansion ratio of convolution layers in an `Op'. There are totally 6 different candidates of kernel size and expansion ratio in each `Op'. }
	\label{fig:choice}
	\vspace{-16pt}
\end{figure}

\textbf{BN Indicator for An Operation.}
When we evaluate the $n$-th ($n=1, \ldots N$) candidate operation $o_{n, l}$ from the $l$-th layer ($l=1, \ldots L$), its BN indicator $S_{o_{n, l}}$ is calculated as follows:

\begin{equation}\label{fuc:op-score}
	\begin{aligned}
	S_{o_{n, l}}= \frac{1}{C}\sum_{c=1}^{C}{\left|{\gamma_{c}^{o_{n,l}}}\right|},
	\end{aligned}
\end{equation}
where ${\gamma_{c}^{o_{n,l}}}$ is the learned parameters of $c$-th channel in chosen candidate operation $o_{n, l}$.
A candidate operation has many CONV, BN, and RELU layers. During the forward-propagation of an operation, features are normalized several times and the final outputs are only determined by the last scaling parameters. 
Thus we utilize only the last BN layer of each building operation as shown by the red box in the right side of Fig.~\ref{fig:choice} to indicate the performance of the candidate operation.
BN indicator needs the last layer of each operation to be a BN layer, so it is not directly applicable to search for models (such as pre-activation ResNes) where BN layers are placed at the beginning of the ops. However, most existing NAS methods apply similar search space as ours, such as recently published works~\cite{EffNet-ICML19-Tan,FBNet-CVPR19-Wu, greedyNAS-cvpr20-you, SPOS-ECCV20-Guo, FairNas-arxiv19-Liang, wan2020fbnetv2}. The BN-indicator can be directly applied to those methods to reduce the computation cost.

\textbf{BN-based Indicator for An Architecture.}
Assuming that there are $L$ search layers in the supernet, we randomly sample the candidate operation $o_{a_l,l}$ from $l$-th layer to construct the subnet architecture $a = [{o_{a_1,1}, ..., o_{a_l,l}, ..., o_{a_{L}, L}}]$. The estimated BN score of the subnet $\mathcal{N}_{a}$ is calculated by

\begin{equation}\label{fuc:subnet-score}
S_{\mathcal{N}_{a}}= \sum\limits_{l = 1}^L {S_{o_{a_{l}, l}}}.
\end{equation}

\textbf{Searching Architecture using BN-based Indicator}
Through calculating the BN score of the subnet, we can estimate the subnet performance without evaluating it on the validation set and the searching stage can be formulated as

\begin{equation}\label{fuc:bn-indicator}
\begin{aligned}
	&a^{*}=\underset{a \in \mathcal{A}}{\operatorname{argmax}} S_{\mathcal{N}_{a}},\\
	&s.t. \ \ FLOPs(a) < Constraint.
\end{aligned}
\end{equation}
For searching an optimal subnet, we randomly sample subnets $\mathcal{N}_{a}$ under the FLOPs constraint and evaluate them based on our BN-based indicator. The optimal subnet is the one with the highest BN score $S_{\mathcal{N}_{a}}$.
Accuracy on validation dataset is a common metric for evaluating subnets in most exsiting NAS methods, while BN-indicator is used in our BN-NAS for evaluating subnets.

\begin{algorithm}[t]
	\caption{BN-based one-shot NAS}
	\label{alg:pipeline}
	{\textbf{Inputs: }}   supernet $\mathcal{N}$ representing Search space $\mathcal{A}$ , Subnet sampling policy on the search space $P(\mathcal{A})$, Training epoch $T$, Sampling subnet number $N_s$ for searching, Training set $\mathcal{T}_{train}$, FLOPs Constraint $F$  \\
	{\textbf{Output:}} Searched Model.\\
	\begin{algorithmic}
		\STATE {\flushleft {\bf{1) Training: }}} 
		\FOR{$epoch \in {0, 1, \hdots, T}$}
		\STATE {\flushleft {Train supernet  through Sampling Policy $P(\mathcal{A})$ on training set $\mathcal{T}_{train}$.}}
		\ENDFOR

		\STATE {\flushleft {\bf{2) Searching: }}} 
		\STATE {\flushleft {Sample $N_s$ subnets under Constraint $F$ and evaluate them based on our BN-based indicator via Eqn.(\ref{fuc:op-score}) and (\ref{fuc:subnet-score}). }}
		\STATE Choose the architecture $a^*$  with highest score as the searching result, Eqn.(\ref{fuc:bn-indicator}).

		\STATE {\flushleft {\bf{3) Retraining: }}} 
		\STATE Train the optimal architecture $a^*$ from scratch on training set $\mathcal{T}_{train}$ and get the trained searched model $M_{a^*}$.
		
	\end{algorithmic}

	{\textbf{Return: }} $M_{a^*}$
\end{algorithm}

\begin{table*}[t]
	\centering
	\caption{Comparison of baseline methods and our method on ImageNet.}
	\centering
	\small
	\begin{tabular}{l|c|c|c|c}
		\hline
		& {SPOS} & {SPOS+Ours} & {FairNAS} & {FairNAS+Ours}   \\
		\hline
		{Top1-ACC (\%)} & 75.73 & 75.67 & 74.07 & 74.12   \\
		\hline
		{FLOPs(M)} & 470 &  470   & 325 & 326   \\
		\hline
		{supernet training epochs} & 100 & 10 & 150 & 15   \\
		\hline
		{supernet training parameters} & All & BN & All  & BN   \\
		\hline
		{Subnets searching cost} & ~1 GPU day & 0.14s on CPU& ~1 GPU day & 0.14s on CPU  \\
		\hline
		{Subnets searching data} &  validation set & None&  validation set & None  \\
		\hline
	\end{tabular}
	\label{tab:main_result} 
	\vspace{-3mm}
\end{table*}

\subsection{Training Only BN Layer}\label{sec:only bn}
Previous work~\cite{TrainBN-arxiv20-Frankle} shows that only training BN layers can still improve the expressive ability of DNN. Since only BN parameters, instead of subnet accuracies, are used during the subnet searching stage, we can only train BN layers instead of the whole supernet in the supernet training stage. 
Specifically, only BN parameters are updated during the back-propagation.

The time required for training supernet by our design is 8\% ($10\% \times 80\% = 8\%$) of the time required by SPOS.  
The reduction in training time comes from two aspects, the fewer epochs (10\%) and training BN only (further 80\%).

1. Fewer epochs. 
The original SPOS method needs to train supernet for 100 epochs while our method needs only 10 epochs (equal to 10\% of original training time).

2. Training BN only. When we train the supernet, we fix the parameters of all convolutional and fully-connected layers. Only the scaling and bias parameters of BN layers are trained through forward-backward propagation. Although the gradients of freezing parameters are calculated during backward propagation, these calculated gradients will not be stored or used for updating. Thus, it will be faster than training all parameters.
Through only training the BN layers of the supernet, the time for training supernet can be saved by about 20\% (equal to 80\%).

\begin{figure}[t]
	\centering
	\includegraphics[width=1\linewidth]{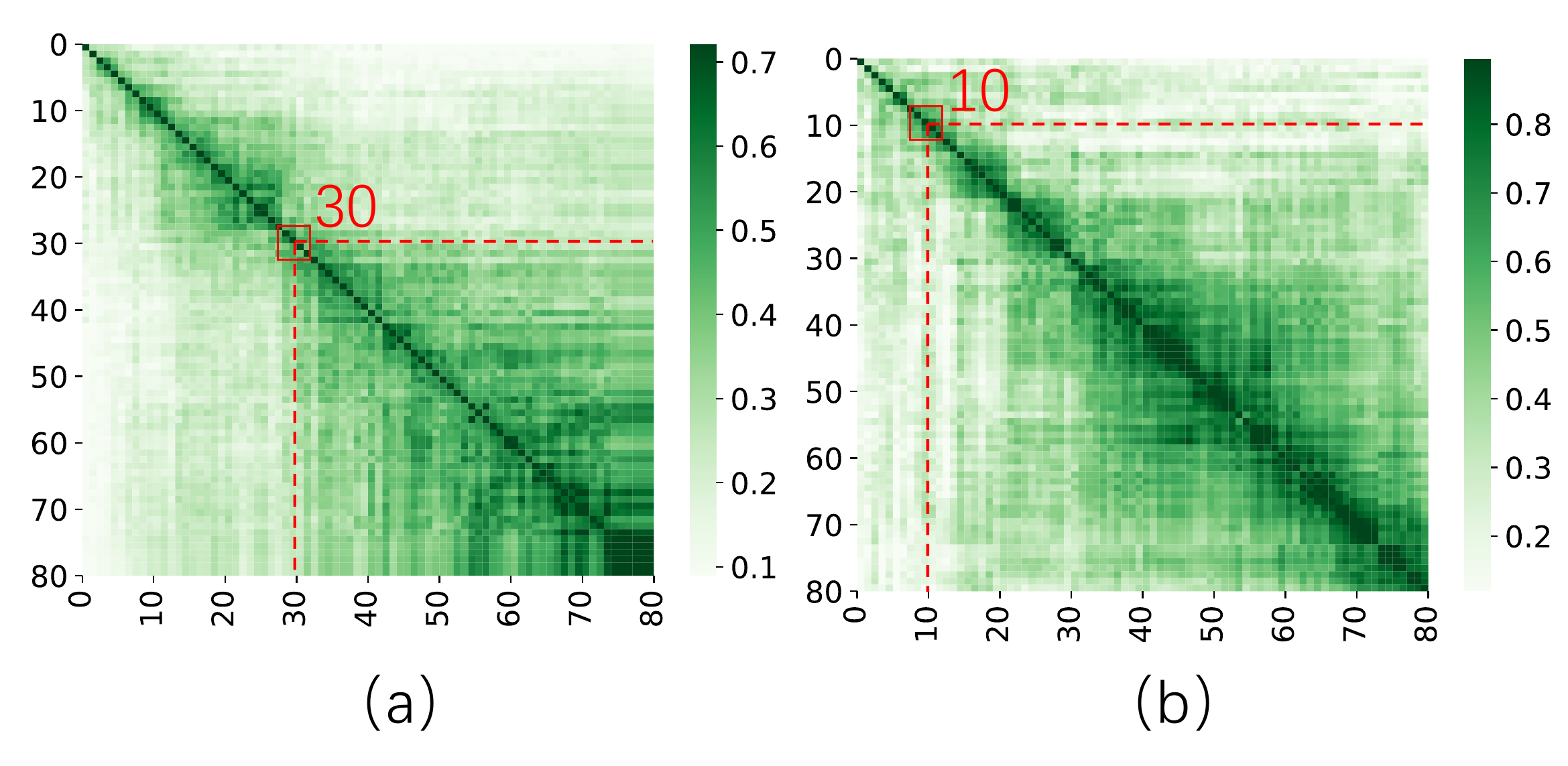}\\
	\caption{Early-bird Characteristic when training the supernet for all parameters (a) and training only BN parameters (b). For better visualization, we normalize the similarity comparison between 0 and 1. The $i,j$-th element in the figure means similarity between $i$-th and $j$-th epoch. Deeper color means higher similarity. We treat the inverted normalized L2-distance as the similarity of two masks from two epochs. 
	Higher value (close to 1) indicates a higher similarity and is highlighted with a darker color. Comparing with training all parameters, training on BN will achieve a faster convergence of BN parameters.}
	\label{fig:early}
	\vspace{-16pt}
\end{figure}

\subsubsection{Analysis on Early-bird Characteristics of BN-based Indicator.}

If accuracy is used for evaluating network architectures, reducing the number of epochs or training only BN would have adverse effect at searching stage. Specifically, the rank from the subnets sampled from the supernet trained in this way would have low correlation with the retrained subnets,
resulting in the subnet sampled from the under-trained supernet unreliable for evaluating the real performance. This adverse effect is observed by the experimental results in Section \ref{subsub:ablation}.
On the other hand, the BN-based Indicator has the early-bird characteristics, which helps us to overcome the potential adverse effect. 

\textbf{Early-bird Characteristics when training all parameters.} Inspired by the inspection of BN parameters for each channel in \cite{EBticket-arxiv19-You}, we investigate the early-bird characteristics of our BN-based indicator. In this setting, all parameters are used. Given the trained supernet, we can evaluate sampled subnets through our proposed BN-based indicator. 
For every epoch during the supernet training, we define a local ranking vector among $N$ candidate operations in the same layer $l$ according to the candidate operation score $S_{o_{1, l}},S_{o_{2, l}}, \ldots, S_{o_{N, l}} $. 
We set the ranking of operation with the highest score as $1$ and operation with the lowest score as rank $N$. By concatenating the $L$ local rank vectors, we can map the trained supernet of an epoch to a ranking vector with size $N \cdot L$. 
For two ranking vectors from two different training epochs of a supernet, we calculate the $L_2$ distance of the two ranking vectors. We visualize the distances among different epochs and find that BN parameters in our supernet training show a similar characteristic as BN in network pruning~\cite{EBticket-arxiv19-You}.  
Fig.~\ref{fig:early}(a) shows the pairwise ranking vector distance matrices (80 $\times$ 80) of the supernet training.
We can find from Fig.~\ref{fig:early}(a) that the similarity between the rank vector at the 30th epoch and the rank vector at the 80th epoch is high. And the rank vector tends to be stable after around the 30th epoch.
This means we can get the optimal architecture information at around the 30th epoch. Therefore, the BN parameters at early training stages are already useful for indicating network performance.

\textbf{Early-bird Characteristics when training BN only.}
When we only train BN layers parameters,  Fig.~\ref{fig:early}(b) shows that the `rank vector' tends to become stable much earlier (at about the 10th epoch) than that for training all parameters in Fig.~\ref{fig:early}(a) (becoming stable at about the 30th epoch).
This means we can use BN-parameters at a very early training stage to find the optimal architecture. We conjecture the reason for faster convergence is that when freezing other parameters and training only BN parameters, the BN parameters try to fit the label with fixed parameters instead of changing parameters, which makes the BN parameters converge much more earlier.
Thus, we can further shorten the training stage through training only BN parameters by one-tenth.  
For clarity, the whole pipeline of our proposed BN-NAS is shown in Algorithm~\ref{alg:pipeline}.

\textbf{Summary of the Early-bird Characteristics.} From the results in Fig.~\ref{fig:early}, we find that: 1) BN-indicator helps the ranking to be stable and facilitates training using fewer epochs; 2) training BN only drives the early-bird characteristics to appear at earlier epochs and facilitates us to train the supernet using much fewer epochs.

\vspace{-1mm}

\section{Experiments}
We first evaluate the BN-based indicator on two basic one-shot NAS methods, including SPOS and FairNAS. Then, we show the ablation experiments to demonstrate the effectiveness of only training BN layers during the supernet training and the early-bird character of the BN indicator. Finally, we verify the transfer ability of the searched model on object detection. Our experiments are tested on NVIDIA GTX 1080Ti GPU with the Pytorch framework.

\subsection{Implementation Details}
\noindent {\bf{Dataset. }}
We evaluate our method on ImageNet~\cite{DengDSLL0-CVPR09-IMAGE}, with 1.28M training samples and 50,000 validation samples. Since we do not need to evaluate our model on the validation set, we only utilize the training samples to train our supernet and the searched subnet.

\noindent {\bf{Search Space. }}
We follow the search space in~\cite{ABS-ECCV20-Hu} , which is composed of MobileNetV2 blocks with kernel size \{3,5,7\}, expansion ratio \{3,6\}. Since our BN-based indicator acts on the BN layer, we do not involve identity operations. We follow the searched depth result of SPOS in~\cite{ABS-ECCV20-Hu} and search other operations during network architecture searching. Since the search space is shrunk, the result of SPOS has been improved. Our experiments are based on the improved version. For the FairNAS search space in~\cite{ABS-ECCV20-Hu}, there are no identity operations.

\noindent {\bf{Hyper-parameters. }} We train the supernet and searched architecture with the same hyper-parameters except the training epoch in all experiments, including only training BN. For network parameters training, we adopt mini-batch Nesterov SGD optimizer with a momentum of 0.9. 
We utilize the learning rate warm-up technique from 0.2 to 0.8 in the first five epochs and adopt cosine annealing learning rate decay from 0.8 to 0. We train the network with a batch size of 1024 and L2 regularization with weight of 1e-4. Besides, the label smoothing is applied with a 0.1 smooth ratio. For baseline supernet training, we train 100 epochs and 150 epochs for SPOS and FairNAS. For our BN supernet training, we use one-tenth of baseline epochs, i.e., 10 epochs and 15 epochs.  For searched architecture retraining, we train the searched architecture from scratch for 240 epochs. For subnet searching, we follow the EA setting in~\cite{SPOS-ECCV20-Guo}. The population size is 50 and max iterations is 20, sampling $N_s = 1000$ subnets under the FLOPs constraint in total.

\begin{table*}[t]
	\centering
	\caption{ImageNet classification results of our method and SOTA. Searching on a small dataset (i.e. CIFAR~\cite{cifar10}) will significantly reduce the search cost. Transferring these methods directly to ImageNet may cause a massive increase in search cost. Some methods even cannot be applied to search on ImageNet due to computational cost. }
	\centering
	\small
	\begin{tabular}{l|c|c|c|c|c|c}
		\hline
		& {Top1-ACC} & params &FLOPs&Search Cost & Search  & Search    \\
		& { (\%)} & (M)&(M)&(GPU days) & method & dataset   \\
		\hline
		\hline
		{ResNet50~\cite{Resnet-CVPR16-He}} & 75.3 & 25.6 &4100 & - & manual & -  \\
		{MobileNetV2(1.4x)~\cite{Mobilev2-CVPR2018-Sandler}} & 74.7 & 6.9 & 585 & - & manual & -  \\
		{ShuffleNetV2(2x)~\cite{Shufflev2-ECCV2018-Ma}} & 74.9 & 7.4 & 591 & - & manual & -  \\
		{EfficientNet-B0(+SE)~\cite{EffNet-ICML19-Tan}} & 76.3 & 5.3 & 390 & - & grid search& ImageNet  \\
		\hline
		{NASNet-A~\cite{NasNet-CVPR18-Zoph}} & 74.0 &5.3 &564 & 2000  & RL & CIFAR  \\
		{AmoebaNet-A~\cite{AmoebaNet-AAAI19-Real}} & 74.5 &5.1 & 555 & 3150  & evolution & CIFAR  \\
		{SNAS(mild)~\cite{SNAS-ICLR19-Xie}} & 72.7 & 4.3 &522 & 1.5  & gradient & CIFAR  \\
		{DARTS~\cite{Darts-ICLR19-Liu}} & 73.3 & 4.7 &574 & 4  & gradient & CIFAR  \\
		{PDARTS~\cite{Pdarts-ICCV19-Chen}} & 75.6 & 4.9 &557 & 0.3  & gradient & CIFAR  \\
		{CARS-G~\cite{CARS-CVPR20-Yang}} & 74.2 & 4.7 &537 & 0.4  & evolution & CIFAR  \\
		\hline
		{ProxylessNAS(GPU)~\cite{Proxyless-ICLR19-Cai}} & 75.1 & 7.1 & 465 & 8.3  & gradient & ImageNet  \\
		{FBNet-C~\cite{FBNet-CVPR19-Wu}} & 74.9 & 5.5 & 375 & 9  & gradient & ImageNet  \\
		{FairNAS~\cite{FairNas-arxiv19-Liang}} & 74.07 & 4.2 & 325 & 16  & evolution & ImageNet  \\
		{SPOS~\cite{SPOS-ECCV20-Guo}} &75.73 & 5.9 & 470 & 11  & evolution & ImageNet  \\
		\hline
		{FairNAS(Ours)} & 74.12 & 3.7 & 326 & 1.2  & evolution & ImageNet  \\
		{SPOS(Ours)} &75.67 & 5.4 & 470 & 0.8  & evolution & ImageNet  \\
		{SPOS(Ours)$+$SE} &76.78 & 7.6 & 473 & 0.8  & evolution & ImageNet  \\
		\hline
	\end{tabular}
	\label{tab:imagenet} 
     \vspace{-18pt} 
\end{table*}

\subsection{Comparison with Baseline Methods}
We compare our method with the baseline methods, SPOS and FairNAS. The comparisons are shown in Table~\ref{tab:main_result}. Our method shortens the NAS process in two stages: supernet training, subnet searching. 

During supernet training, benefited from the early-bird characteristic of the proposed BN-based indicator, we significantly reduce the training epochs of supernet, from 100 to 10 for SPOS and  150 to 15 for FairNAS.  Besides, we only train BN parameters instead of all parameters, which has an additional 20\% speedup for supernet training. 

During subnet searching, baseline methods adopt EA algorithm to sample 1000 subnets and evaluate each on the validation set.  The cost of evaluating 1000 subnets is about 1 GPU day. Our method also samples 1000 subnets but utilize the BN-indicator for subnet evaluation, greatly reducing the evaluation cost from 1 GPU day to 0.14s on CPU. 

Overall, our method accelerates the one-shot NAS method about ten times compared with baseline methods while the performance is still comparable.

\subsection{Comparison with State-of-The-Art Methods}
We compare our method with state-of-the-art (SOTA) methods as the Table~\ref{tab:imagenet} shows. 
Compared with the manual designed networks, our searched model based on SPOS achieves higher performance with fewer FLOPs. Comparing with SOTA NAS methods, regardless of the gradient-based method (Proxyless) or evolution-based (CARS-G) method, our searched model also performs better with the fewer or similar FLOPs. 

For the search cost, our method needs comparable search costs with methods searching on CIFAR and transferring the architecture to ImageNet. For the methods directly searching the architecture on ImageNet, our method requires less than one-tenth of the search cost. Compared with EfficientNet-B0, the grid search in EfficientNet-B0 needs to train plenty of models fully on ImageNet, which has much more search cost than many evolution methods.

\subsection{Detection}
We further validate the transfer ability of our BN-NAS on object detection. We utilize our BN-NAS(SPOS) pretrained on ImageNet as the feature extractor and follow the training setting in EfficientDet~\cite{EffDet-CVPR20-Tan} and use the same detection head as~\cite{EffDet-CVPR20-Tan}. With similar FLOPs as ~\cite{EffDet-CVPR20-Tan}, our searched model achieves comparable performance. Comparing with other manual designed light networks, our searched model achieves better performance with much fewer FLOPs. The model searched with our method has a good transfer ability.

\begin{table}[H]
	\centering
	\caption{Performance of our searched model and some SOTA light models on COCO dataset. Our methods achieve comparable performance with EfficientDet-D0 with much less search cost.}
	\centering
	\begin{tabular}{l|c|c}
		\hline
		backbone &  {FLOPs (B)} & {mAP}   \\
		\hline
		{ShuffleNetv2} & 14 & 27.6    \\
		\hline
		{MobileNetV2}  & 8& 31.7    \\
		\hline
		{ResNet18} & 21 & 32.2    \\
		\hline
		{EfficientDet-D0}  &2.5 & 33.46  \\
		\hline
		{Ours} & 2.7   &  33.32 \\
		\hline
	\end{tabular}
	\label{tab:detection} 
\end{table}

\subsection{Ablation Experiments}\label{subsub:ablation}
In this section, we design experiments to show the effectiveness of BN-indicator (in Section~\ref{subsub:indicators}) and show the similar correlation relationship between the BN-indicator score and retrain accuracy compared with SPOS~\cite{SPOS-ECCV20-Guo} (in Section~\ref{subsub:correlation}). More experiments about different initialization methods are in supplementary materials.

\vspace{-4mm}
\subsubsection{Indicators}\label{subsub:indicators}
\begin{figure}[t]
	\centering
	\includegraphics[width=1\linewidth]{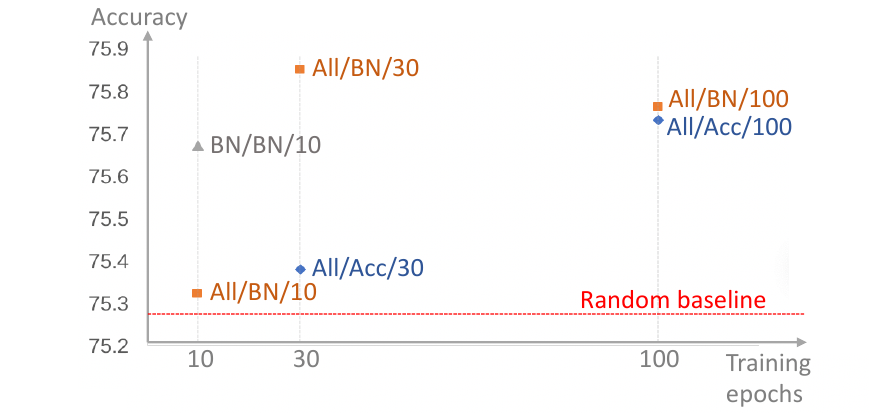}\\
	\caption{The accuracy of searched architectures with different training settings on ImageNet. `All/BN/k' means training all parameters of the supernet for k epochs and using BN-indicator to find optimal subnets. In `All/Acc/k', Acc-indicator is used instead of BN-indicator. In `BN/BN/k', only BN parameters of the supernet is trained. Red dotted line shows accuracy of random baseline.}
	\label{fig:abaepo}
	\vspace{-12pt}
\end{figure}

Our BN-indicator is used for evaluating subnet during the searching process. Most existing NAS methods utilize the model accuracy on validation dataset to evaluate subnet, which is denoted as Acc-indicator here. Besides, we also randomly sampled five subnets from the supernet and choose the subnet with the highest accuracy as the random basline, as shown in red dotted line in Fig.~\ref{fig:abaepo}.

{\flushleft\textbf{Training all parameters for 100 epochs.}}
We train the supernet for 100 epochs and search the subnet based on BN-indicator (`All/BN/100' in Fig.~\ref{fig:abaepo}) and Acc-indicator (`All/Acc/100' in Fig.~\ref{fig:abaepo}). The searched subnets with these two indicators perform similarly, showing that BN-indicator is on par with accuracy as the Acc-indicator when all parameters are trained for enough training epochs. However, both training the SurperNet for 100 epochs and evaluating model accuracy leads to much computation cost.

\vspace{-2mm}
{\flushleft\textbf{Training all parameters for 30 epochs.}}
We reduce the training epochs of supernet from 100 to 30 and test the performance of two indicator under this setting, as shown in Fig.~\ref{fig:abaepo}. 
With less training epochs, the performance of subnets from BN-indicator (`All/BN/30' in Fig.~\ref{fig:abaepo}) perform better than those from Acc-indicator (`All/Acc/30' in Fig.~\ref{fig:abaepo}). It shows that longer training epoch of supernet is essential for Acc-indicator but not important for BN-indicator. 
The searched model using Acc-indicator (`All/Acc/30' in Fig.~\ref{fig:abaepo}) performs only better than random baseline (red dotted line in Fig.~\ref{fig:abaepo}) by a small margin of 0.1\%. The performance drop on the low correlation is caused by the incomplete training. The accuracy of subnets from supernet which only trained for 30 epochs cannot represent the retraining accuracy precisely, causing the searched model performs near random baseline. {On the other hand, the BN parameters of the supernet shows Early-bird characteristics. As shown in Fig.~\ref{fig:early}(a), it shows stronger correlation between the supernets trained for 30 epochs and 80 epochs, considering only their BN values. The Early-bard characteristics explains well why the proposed BN-indicator still keeps good performance for 30 epochs. } 

\vspace{-2mm}
{\flushleft\textbf{Training all parameters for 10 epochs.}}
When further reducing the training epochs from 30 to 10, our BN-indicator cannot keep good performance, as shown in Fig.~\ref{fig:abaepo}. The reason is that the BN parameters are not trained well at 10 epoch during the supernet training if all parameters are trained, as show in Fig.~\ref{fig:early} (a). Training all parameters leads to inconsistent convolution parameters at different training epochs, hindering the BN parameters from converging faster. Inspired by this insight, we try to reduce the required training epoch of supernet by only training the BN parameters. As shown in Fig.~\ref{fig:early} (b), the Early-bird characteristic appears even earlier at about 10 epoch when we only train BN parameters in the supernet. By training BN-only, our BN-NAS returns to its excellent performance during subnet searching, as shown by `BN/BN/10'  in Fig.~\ref{fig:abaepo}.

\begin{figure}[t]
	\centering
	\includegraphics[width=1\linewidth]{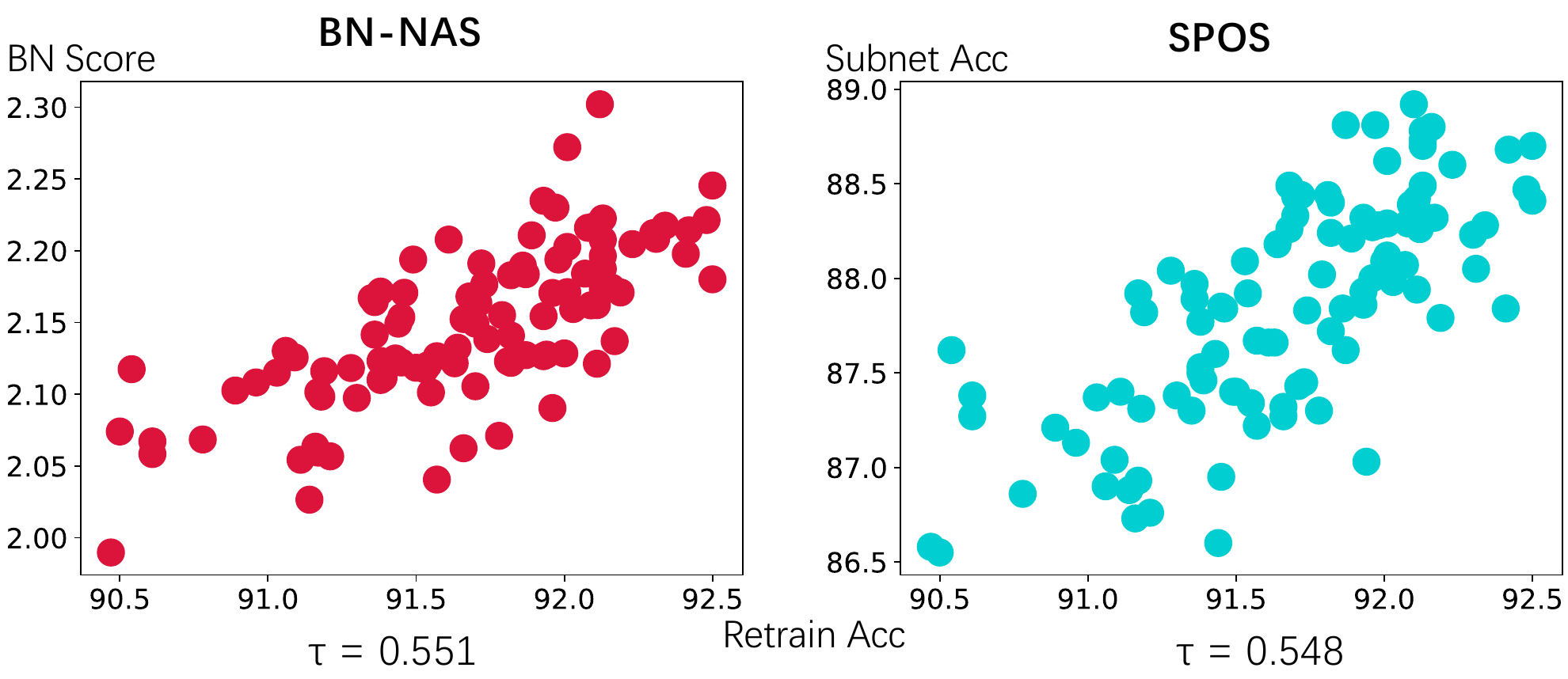}\\
	\caption{Model correlations for Ours and SPOS.}
	\label{fig:corr}
  \vspace{-12pt}
\end{figure}

\vspace{-3mm}
\subsubsection{Correlation with retrain accuracy}\label{subsub:correlation}
\vspace{-1mm}
For one-shot NAS methods, a well-known problem is the low performance consistency of different subnets. Indicator plays an important role to keep high performance consistency in one-shot NAS. To evaluate the effectiveness of our proposed BN-indicator, we conduct the correlation experiments on CIFAR10 dataset. We follow the search space in~\cite{ABS-ECCV20-Hu} and train the supernet for 600 epochs. Then we randomly sample 100 architectures and retrain them from scratch. We utilize Kendall Tau $\tau$ metric to show the correlation between the BN-score obtained by Eqn.(\ref{fuc:subnet-score}) and the retrain accuracy of sampled models. We also show the correlation between validation accuracy and retrain accuracy of models which are sampled based on Acc-indicator (SPOS).
As shown in Fig.~\ref{fig:corr}, our method achieves similar Kendall Tau $\tau$ as SPOS, which means the proposed BN-indicator has a good indication ability as the Acc-indicator used in SPOS. Our experiments in achieving similar accuracy of the searched model also support this conclusion.

\vspace{-2mm}
\section{Conclusions}
NAS has greatly boosted the SOTA methods with deep networks in computer vision.  However, existing NAS methods are time-consuming. We propose a novel BN-based indicator to efficiently evaluate the performance of subnets selected from the supernet, drastically accelerating the searching process for NAS. Thanks to the Early-bird character, we can train the supernet by only training the BN layers, further reducing supernet training time. Our extensive experiments validate that the proposed BN-NAS can decrease the whole time consumption for one-shot NAS. 

{\flushleft\textbf{Acknowledgements}}
This work was supported by the Australian Research Council Grant DP200103223, FT210100228, and Australian Medical Research Future Fund MRFAI000085.

{\small
\bibliographystyle{ieee_fullname}
\bibliography{egbib}

\begin{thebibliography}{10}\itemsep=-1pt

\bibitem{RLnas-ICML17-BelloZVL}
Irwan Bello, Barret Zoph, Vijay Vasudevan, and Quoc~V. Le.
\newblock Neural optimizer search with reinforcement learning.
\newblock In {\em ICML}, 2017.

\bibitem{Proxyless-ICLR19-Cai}
Han Cai, Ligeng Zhu, and Song Han.
\newblock Proxylessnas: Direct neural architecture search on target task and
  hardware.
\newblock In {\em ICLR}, 2019.

\bibitem{chen2021glit}
Boyu Chen, Peixia Li, Chuming Li, Baopu Li, Lei Bai, Chen Lin, Ming Sun, Junjie
  Yan, and Wanli Ouyang.
\newblock Glit: Neural architecture search for global and local image
  transformer.
\newblock In {\em ICCV}, 2021.

\bibitem{DeepLabv3+NIPS18-Chen}
Liang{-}Chieh Chen, Maxwell~D. Collins, Yukun Zhu, George Papandreou, Barret
  Zoph, Florian Schroff, Hartwig Adam, and Jonathon Shlens.
\newblock Searching for efficient multi-scale architectures for dense image
  prediction.
\newblock In {\em NeurIPS}, 2018.

\bibitem{Pdarts-ICCV19-Chen}
Xin Chen, Lingxi Xie, Jun Wu, and Qi Tian.
\newblock Progressive differentiable architecture search: Bridging the depth
  gap between search and evaluation.
\newblock In {\em ICCV}, 2019.

\bibitem{DetNAS-NIPS19-Chen}
Yukang Chen, Tong Yang, Xiangyu Zhang, Gaofeng Meng, Xinyu Xiao, and Jian Sun.
\newblock Detnas: Backbone search for object detection.
\newblock In {\em NeurIPS}, 2019.

\bibitem{FairNas-arxiv19-Liang}
Xiangxiang Chu, Bo Zhang, Ruijun Xu, and Jixiang Li.
\newblock Fairnas: Rethinking evaluation fairness of weight sharing neural
  architecture search.
\newblock {\em CoRR}, abs/1907.01845, 2019.

\bibitem{ci2020evolving}
Yuanzheng Ci, Chen Lin, Ming Sun, Boyu Chen, Hongwen Zhang, and Wanli Ouyang.
\newblock Evolving search space for neural architecture search.
\newblock In {\em ICCV}, 2021.

\bibitem{DengDSLL0-CVPR09-IMAGE}
Jia Deng, Wei Dong, Richard Socher, Li{-}Jia Li, Kai Li, and Fei{-}Fei Li.
\newblock Imagenet: {A} large-scale hierarchical image database.
\newblock In {\em CVPR}, 2009.

\bibitem{TrainBN-arxiv20-Frankle}
Jonathan Frankle, David~J. Schwab, and Ari~S. Morcos.
\newblock Training batchnorm and only batchnorm: On the expressive power of
  random features in cnns.
\newblock {\em CoRR}, abs/2003.00152, 2020.

\bibitem{SPOS-ECCV20-Guo}
Zichao Guo, Xiangyu Zhang, Haoyuan Mu, Wen Heng, Zechun Liu, Yichen Wei, and
  Jian Sun.
\newblock Single path one-shot neural architecture search with uniform
  sampling.
\newblock In {\em ECCV}, 2020.

\bibitem{Resnet-CVPR16-He}
Kaiming He, Xiangyu Zhang, Shaoqing Ren, and Jian Sun.
\newblock Deep residual learning for image recognition.
\newblock In {\em CVPR}, 2016.

\bibitem{ABS-ECCV20-Hu}
Yiming Hu, Yuding Liang, Zichao Guo, Ruosi Wan, Xiangyu Zhang, Yichen Wei,
  Qingyi Gu, and Jian Sun.
\newblock Angle-based search space shrinking for neural architecture search.
\newblock {\em CoRR}, abs/2004.13431, 2020.

\bibitem{BN-ICML15-Ioffe}
Sergey Ioffe and Christian Szegedy.
\newblock Batch normalization: Accelerating deep network training by reducing
  internal covariate shift.
\newblock In {\em ICML}, 2015.

\bibitem{OP-aware-pruning-arxiv20-Kang}
Minsoo Kang and Bohyung Han.
\newblock Operation-aware soft channel pruning using differentiable masks.
\newblock {\em CoRR}, abs/2007.03938, 2020.

\bibitem{cifar10}
Alex Krizhevsky, Geoffrey Hinton, et~al.
\newblock Learning multiple layers of features from tiny images.
\newblock 2009.

\bibitem{li2020improving}
Xiang Li, Chen Lin, Chuming Li, Ming Sun, Wei Wu, Junjie Yan, and Wanli Ouyang.
\newblock Improving one-shot nas by suppressing the posterior fading.
\newblock In {\em CVPR}, 2020.

\bibitem{CRNAS-ICLR19-Liang}
Feng Liang, Chen Lin, Ronghao Guo, Ming Sun, Wei Wu, Junjie Yan, and Wanli
  Ouyang.
\newblock Computation reallocation for object detection.
\newblock In {\em ICLR}, 2020.

\bibitem{AutoDeepLab-CVPR2019-Liu}
Chenxi Liu, Liang{-}Chieh Chen, Florian Schroff, Hartwig Adam, Wei Hua, Alan~L.
  Yuille, and Fei{-}Fei Li.
\newblock Auto-deeplab: Hierarchical neural architecture search for semantic
  image segmentation.
\newblock In {\em CVPR}, 2019.

\bibitem{Darts-ICLR19-Liu}
Hanxiao Liu, Karen Simonyan, and Yiming Yang.
\newblock {DARTS:} differentiable architecture search.
\newblock In {\em ICLR}, 2019.

\bibitem{liu2021inception}
Jie Liu, Chuming Li, Feng Liang, Chen Lin, Ming Sun, Junjie Yan, Wanli Ouyang,
  and Dong Xu.
\newblock Inception convolution with efficient dilation search.
\newblock In {\em CVPR}, 2021.

\bibitem{NetworkSliming-ICCV17-Liu}
Zhuang Liu, Jianguo Li, Zhiqiang Shen, Gao Huang, Shoumeng Yan, and Changshui
  Zhang.
\newblock Learning efficient convolutional networks through network slimming.
\newblock In {\em ICCV}, 2017.

\bibitem{Shufflev2-ECCV2018-Ma}
Ningning Ma, Xiangyu Zhang, Hai{-}Tao Zheng, and Jian Sun.
\newblock Shufflenet {V2:} practical guidelines for efficient {CNN}
  architecture design.
\newblock In {\em ECCV}, 2018.

\bibitem{ENAS-ICML18-Pham}
Hieu Pham, Melody~Y. Guan, Barret Zoph, Quoc~V. Le, and Jeff Dean.
\newblock Efficient neural architecture search via parameter sharing.
\newblock In {\em ICML}, 2018.

\bibitem{AmoebaNet-AAAI19-Real}
Esteban Real, Alok Aggarwal, Yanping Huang, and Quoc~V. Le.
\newblock Regularized evolution for image classifier architecture search.
\newblock In {\em AAAI}, 2019.

\bibitem{Mobilev2-CVPR2018-Sandler}
Mark Sandler, Andrew~G. Howard, Menglong Zhu, Andrey Zhmoginov, and
  Liang{-}Chieh Chen.
\newblock Mobilenetv2: Inverted residuals and linear bottlenecks.
\newblock In {\em CVPR}, 2018.

\bibitem{EffNet-ICML19-Tan}
Mingxing Tan and Quoc~V. Le.
\newblock Efficientnet: Rethinking model scaling for convolutional neural
  networks.
\newblock In {\em ICML}, 2019.

\bibitem{EffDet-CVPR20-Tan}
Mingxing Tan, Ruoming Pang, and Quoc~V. Le.
\newblock Efficientdet: Scalable and efficient object detection.
\newblock In {\em CVPR}, 2020.

\bibitem{wan2020fbnetv2}
Alvin Wan, Xiaoliang Dai, Peizhao Zhang, Zijian He, Yuandong Tian, Saining Xie,
  Bichen Wu, Matthew Yu, Tao Xu, Kan Chen, et~al.
\newblock Fbnetv2: Differentiable neural architecture search for spatial and
  channel dimensions.
\newblock In {\em CVPR}, 2020.

\bibitem{FBNet-CVPR19-Wu}
Bichen Wu, Xiaoliang Dai, Peizhao Zhang, Yanghan Wang, Fei Sun, Yiming Wu,
  Yuandong Tian, Peter Vajda, Yangqing Jia, and Kurt Keutzer.
\newblock Fbnet: Hardware-aware efficient convnet design via differentiable
  neural architecture search.
\newblock In {\em CVPR}, 2019.

\bibitem{SNAS-ICLR19-Xie}
Sirui Xie, Hehui Zheng, Chunxiao Liu, and Liang Lin.
\newblock {SNAS:} stochastic neural architecture search.
\newblock In {\em ICLR}, 2019.

\bibitem{CARS-CVPR20-Yang}
Zhaohui Yang, Yunhe Wang, Xinghao Chen, Boxin Shi, Chao Xu, Chunjing Xu, Qi
  Tian, and Chang Xu.
\newblock {CARS:} continuous evolution for efficient neural architecture
  search.
\newblock In {\em CVPR}, 2020.

\bibitem{EBticket-arxiv19-You}
Haoran You, Chaojian Li, Pengfei Xu, Yonggan Fu, Yue Wang, Xiaohan Chen,
  Yingyan Lin, Zhangyang Wang, and Richard~G. Baraniuk.
\newblock Drawing early-bird tickets: Towards more efficient training of deep
  networks.
\newblock {\em CoRR}, abs/1909.11957, 2019.

\bibitem{greedyNAS-cvpr20-you}
Shan You, Tao Huang, Mingmin Yang, Fei Wang, Chen Qian, and Changshui Zhang.
\newblock Greedynas: Towards fast one-shot {NAS} with greedy supernet.
\newblock In {\em CVPR}, 2020.

\bibitem{EcoNAS-CVPR20-Zhou}
Dongzhan Zhou, Xinchi Zhou, Wenwei Zhang, Chen~Change Loy, Shuai Yi, Xuesen
  Zhang, and Wanli Ouyang.
\newblock Econas: Finding proxies for economical neural architecture search.
\newblock In {\em CVPR}, 2020.

\bibitem{NasNet-CVPR18-Zoph}
Barret Zoph, Vijay Vasudevan, Jonathon Shlens, and Quoc~V. Le.
\newblock Learning transferable architectures for scalable image recognition.
\newblock In {\em CVPR}, 2018.

\end{thebibliography}
}

\end{document}